%% file: main.tex
\def\year{2020}\relax
\documentclass[letterpaper]{article} 
\usepackage{aaai20}  
\usepackage{times}  
\usepackage{helvet} 
\usepackage{courier}  
\usepackage[hyphens]{url}  
\usepackage{graphicx} 
\usepackage{graphicx}  
\usepackage{subfigure}

\title{Impact of Explanation on Trust of a Novel Mobile Robot}
\author{Stephanie Rosenthal and Elizabeth J. Carter\\ 
Carnegie Mellon University\\
Pittsburgh PA 15213\\
\{rosenthal,lizcarter\}@cmu.edu 
}

\begin{document}

\maketitle

\begin{abstract}
One challenge with introducing robots into novel environments is misalignment between supervisor expectations and reality, which can greatly affect a user's trust and continued use of the robot. We performed an experiment to test whether the presence of an explanation of expected robot behavior affected a supervisor's trust in an autonomous robot. We measured trust both subjectively through surveys and objectively through a dual-task experiment design to capture supervisors' neglect tolerance (i.e., their willingness to perform their own task while the robot is acting autonomously). Our objective results show that explanations can help counteract the novelty effect of seeing a new robot perform in an unknown environment. Participants who received an explanation of the robot's behavior were more likely to focus on their own task at the risk of neglecting their robot supervision task during the first trials of the robot's behavior compared to those who did not receive an explanation. However, this effect diminished after seeing multiple trials, and participants who received explanations were equally trusting of the robot's behavior as those who did not receive explanations. Interestingly, participants were not able to identify their own changes in trust through their survey responses, demonstrating that the dual-task design measured subtler changes in a supervisor's trust.
\end{abstract}

\input{Introduction}

\input{Method.tex}
\input{Results.tex}

\input{Discussion.tex}

\input{Conclusion.tex}

\bibliographystyle{aaai}
\bibliography{trust}

\end{document}

%% file: Introduction.tex
\section{Introduction and Related Work}
As we introduce robots that perform tasks into our environments, the people who live and work around the robots will be expected to maintain their own productivity while largely ignoring the robots as they move around and complete jobs. While this pattern of behavior around robots can be expected to develop over time, the introduction of a new robot is frequently disruptive to people in its environment in several ways. First, people are uncertain of a robot's autonomous behaviors when it is first introduced. People for whom a robot is novel are typically observed testing the robot's abilities (e.g., \cite{gockley2005,bohus2014}) and monitoring robot behavior in the environment rather than executing their own tasks (e.g., \cite{burgard1998interactive,thrun1999minerva,kanda2010communication,rosenthal2012mobile}). Additionally, even people who understand basic behaviors of robots often must intervene to help robots overcome failures or errors in their autonomy \cite{de2006comprehensive}. Failures impact both human productivity and their trust in the robot's behavior \cite{desai2012effects}.

One proposed technique to create appropriate user expectations \cite{tolmeijer2020trusttaxonomy} and overcome the challenges of human uncertainty and mistrust for different types of intelligent systems is to provide feedback and explanations to users (e.g., \cite{lim2009and,ribeiro2016should,desai2013impact,abdul2018trends}).  \citeauthor{bussone15} \shortcite{bussone15} found that explanations of machine learning predictions align user's mental models such that they increase their trust and reliance on the predictions. Recent work has extended the idea of explainability to robot decision processes to help people understand, for example, why a robot performs an action based on its policy \cite{hayes2017improving} or its reward function \cite{sukkerd2018towards}, or to summarize the robot's recent actions for different people and/or purposes \cite{rosenthal2016verbalization}. While explanation algorithms have been successfully compared to human explanations, little has been done to understand how explanations impact trust of autonomous mobile robots.

Most commonly, researchers use subjective surveys to measure trust on binary \cite{hall1996binarytrust}, ordinal \cite{muir1989operators}, and continuous \cite{lee1992trust} scales. These scales can be measured one time or many times to build up metrics such as the Area Under the Trust Curve (AUTC) \cite{desai2012effects} and Trust of Entirety \cite{yang2017evaluating}. Objective measures of trust have also been proposed, including neglect tolerance, which we use in this work. The neglect time of a robot is the mean amount of time that the robot can function with task performance above a certain threshold without human intervention, and neglect tolerance is the overall measure of robot autonomy when neglected \cite{goodrich2003seven}. Neglect tolerance in particular is an important objective measure because autonomous operation is a primary goal of robotics research and development. For a user, it is a key contributor to the amount of trust that can be placed in a robot. The more a user can neglect instead of attend to a robot, the more they can focus on other tasks. 

Towards the goal of measuring the effects of explanations on both subjective and objective trust, we designed a dual-task experiment in which participants were asked to allocate their attention to a robot's behavior and a video game. Participants were asked to play a basic video game while monitoring a robot as it navigated a path across a large grid map. They were directed to note if the robot entered certain squares on the grid while also playing the game to their maximum ability. Some participants also received an explanation of how the robot would move around the map and which squares it would try to avoid, and some did not. 
We used video game performance as a proxy for measuring neglect tolerance, trust and reliance. Slower and less successful gameplay indicated that more attention was diverted to the robot, which in turn implied less trust and reliance on the autonomy. Additionally, we used questionnaires to examine subjective ratings of trust in the robot across navigation conditions. We hypothesized that receiving an explanation of the robot's behavior would increase trust as measured by the time allocated to gameplay versus supervising the robot as well as subjective trust ratings.

Our results show that our dual-task experiment was able to measure differences in trust. Key press count and key press rate for the game were both slower when the robot made errors by entering target squares compared to when it did not, indicating that the participants spent more time monitoring the robot during periods when an error occurred. Participants' time to report an error did not drop, indicating that participants traded off gameplay performance (key presses) in order to monitor the robot rather than miss an opportunity to report an error. In contrast, our surveys did not show any differences in trust when the robot made errors compared to when they did not. This result indicates that our dual-task experiment measured subtle changes in trust that the survey results could not identify.

The presence of an explanation additionally affected participant trust behaviors in the first two of three trials. The results show that participants who had received an explanation of the robot's behavior had higher key press counts and a lower number of game losses during times when the robot made errors in the first two trials compared to the last trial. This result indicates that participants initially were able to focus on the game more because the explanation gave them information about the robot's behavior. However, by the third trial, all participants understood the robot's behavior and the effect of explanation was no longer significant. We conclude that an explanation can help counteract the novelty effect an unfamiliar robot by improving user trust.

%% file: Method.tex
\begin{figure*}[t]
    \centering
    \subfigure[Egypt map]{
    \label{fig:egypt}
    \includegraphics[width=0.27\textwidth]{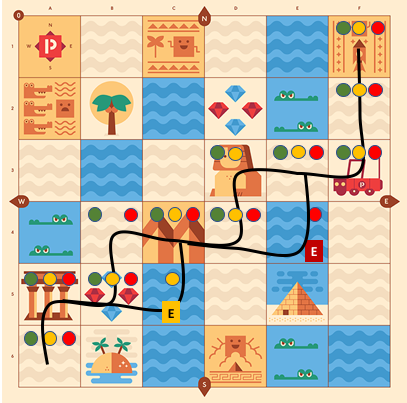}
    }
    \subfigure[Space map]{
        \label{fig:space}
    \includegraphics[width=0.27\textwidth]{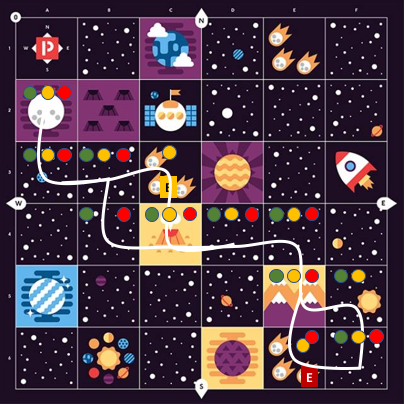}
    }
    \subfigure[Street map]{
        \label{fig:street}
    \includegraphics[width=0.27\textwidth]{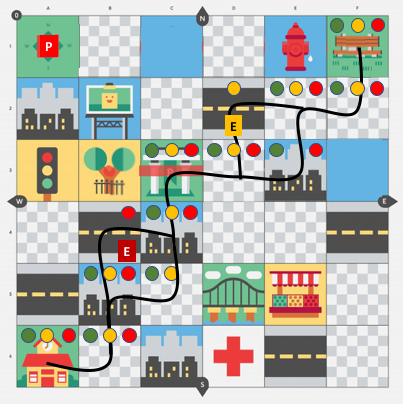}
    }
    \caption{Participants were each asked to monitor three robots executing tasks (one per map). They were each randomly assigned to a condition order in which they saw a path with no error (indicated with green circles), early error (yellow circles), or late error (red circles) for 6 total possible condition combinations.}
        \label{fig:maps}
\end{figure*}

\section{Experiment Method}

In order to measure the effect of explanations on trust, we performed a dual-task experiment in which we asked participants to simultaneously play an online game while they monitored each of three robots as they navigated their large grid maps. Half of our participants also received a brief explanation before the robots executed their tasks of what squares on the map each of the robots was programmed to avoid, while half were given no explanation about the robot's programming. For neglect tolerance, we measured the time that participants spent playing our game as well as the time it took them to report that the robot was making an error (i.e., entering a target type of square) during its execution. Between each robot execution, participants also completed a questionnaire about aspects of trust they had in that particular robot during that trial. We evaluated the differences in each of our dependent measures to understand the how the tasks affected participant trust through the study.

\subsection{Autonomous Robot Navigation Setup}
We used a Cozmo robot from Anki for this experiment. Cozmo is a small robot that can move across flat surfaces on two treads, and it has an associated software development kit (SDK) allows for programming and executing a large range of behaviors, including navigation. 

We programmed Cozmo to navigate Adventure Maps from the Cubetto playset available from Primo Toys. These maps are large, colorful grids with six rows and columns of icons and measure approximately 1 by 1 meter each. Each square in the grid fell into one of two categories: background patterns or icons. For each category, there were multiple examples: a background on an Egypt-themed map could be water or sand, and an icon on a space-themed map included a rocketship. We used the maps from the Big City (referred to as the Street map), Deep Space (Space map), and Ancient Egypt (Egypt map) Adventure Packs mounted on foam core poster board for stability (Figure~\ref{fig:maps}). 

Three paths were chosen for each map: one that did not enter a particular type of square on the map (\textit{no error}), one that entered that square type at the beginning of the path (\textit{early error}), and one that entered that square type at the end of the path (\textit{late error}). The Egypt map had ten water squares that were defined as errors. In the Space map, three comet squares were identified as errors. The Street map contained five error squares that looked like streets. The paths and indicated errors are shown in Figure~\ref{fig:maps}, and these are referred to as the Error Finding conditions. 

Cozmo navigated the paths in an open loop as it was not actively sensing its location on the maps. Cozmo's paths were found to be very consistent in terms of the robot staying in the required squares throughout the study. The experimenter could select a map and a specific path at the beginning of each trial. All of the robot's motions were logged and timestamped in a file labeled by the participant number and their error condition order.

We told participants that there were three distinct robots indicated by different colored tapes on their backs in order to reduce potential confusion about whether the robots were using the same algorithms. However, there was only one actual robot used for consistency in navigation.

\subsection{Participant Tasks}

Participants were asked to simultaneously supervise the robots as they navigated the maps and maximize their scores in an online game of Snake. 

\subsubsection{Supervisory Task}

Participants were asked to indicate by a button press when the robot entered the indicated type of error square (i.e., whether/when it enters water for the Egypt map, a comet square on the Space map, and a street square on the Street map). This task required them to maintain some knowledge about where the robot was located on the map and where the potential error squares were located, typically by occasionally watching the robot's behavior. 




\subsubsection{Snake Game Task}

In order to simulate a real-world scenario in which the human supervisor of a robot would need to perform other tasks at the same time (including, perhaps, supervising multiple robots or performing their own task), we created another responsibility for our participants. While the robot was navigating its path, participants were provided with a laptop on which to play a web-based, single-player game of Snake. The goal of Snake is to direct a moving chain of yellow squares (the snake) around the screen using the arrow keys and collect as many additional red food squares as possible by aiming the snake directly at them and bumping them with the head of the snake (Figure~\ref{fig:snake}). We asked participants to maximize their score in the game by collecting as many food pieces as possible without hitting one of the outer walls (in this case, red squares positioned along the edges of the gameplay window) or accidentally hitting the snake body with the head (which becomes more difficult as the snake becomes longer). In these cases, the snake dies and participants start over. Participants were not able to pause the game, so they had to make tradeoffs in their gameplay in order to successfully monitor the robot.

By hosting the Snake game on a website, we were able to collect data about every button press made, the score at any time, the duration of each game, and whether participants had to restart the game due to the snake hitting obstacles or itself. These data were collected on every event and measured to the millisecond. We used these logs to measure differences in the rate and count of key presses and the number of obstacles hit (game deaths) across trials. The degree to which participants were attending to the game versus visually inspecting the robot's progress and monitoring its errors should be apparent in gameplay slowdown and/or increases in obstacles hit when participants are not watching the Snake's motion.

\subsection{Explanation Condition}
The key between-subjects variable for this experiment was the explanation provided to the participants about Cozmo's navigation behavior. There are many possible explanations we could have provided, including summaries of the path the robot would take and the policy in each grid square. However, we chose a short explanation that followed a similar pattern found in prior work \cite{li2017trajectories} in which preferences for particular squares were noted. This brief explanation format was developed to be easy to understand and recall while not inducing the attribution of goals and mental states to the robot. In this experiment, only a single square type was avoided, so it was simple and concise to provide participants with this information. 

Half of the participants (No Explanation condition) were only told the map description (Egypt, Space, Street) and to press the button if Cozmo entered one of the error squares (water, comets, or street). For example:
\begin{quote}
    ``This Cozmo navigates the space map. Hit the button if Cozmo hits the comets.''
\end{quote}
For the other half of the participants (Explanation condition), an additional explanation was provided to explain why the participants were being directed to hit the button if the Cozmo entered an error square: to report the mistake.
\begin{quote}
    ``This Cozmo navigates the space map and is programmed to avoid the comets. Hit the button if Cozmo hits the comets anyway.''
\end{quote} 


\subsection{Study Design}

\subsubsection{Experimental Setup}
\begin{figure*}[t]
\centering
\subfigure[Experimental Setup]{
\includegraphics[width=0.25\textwidth]{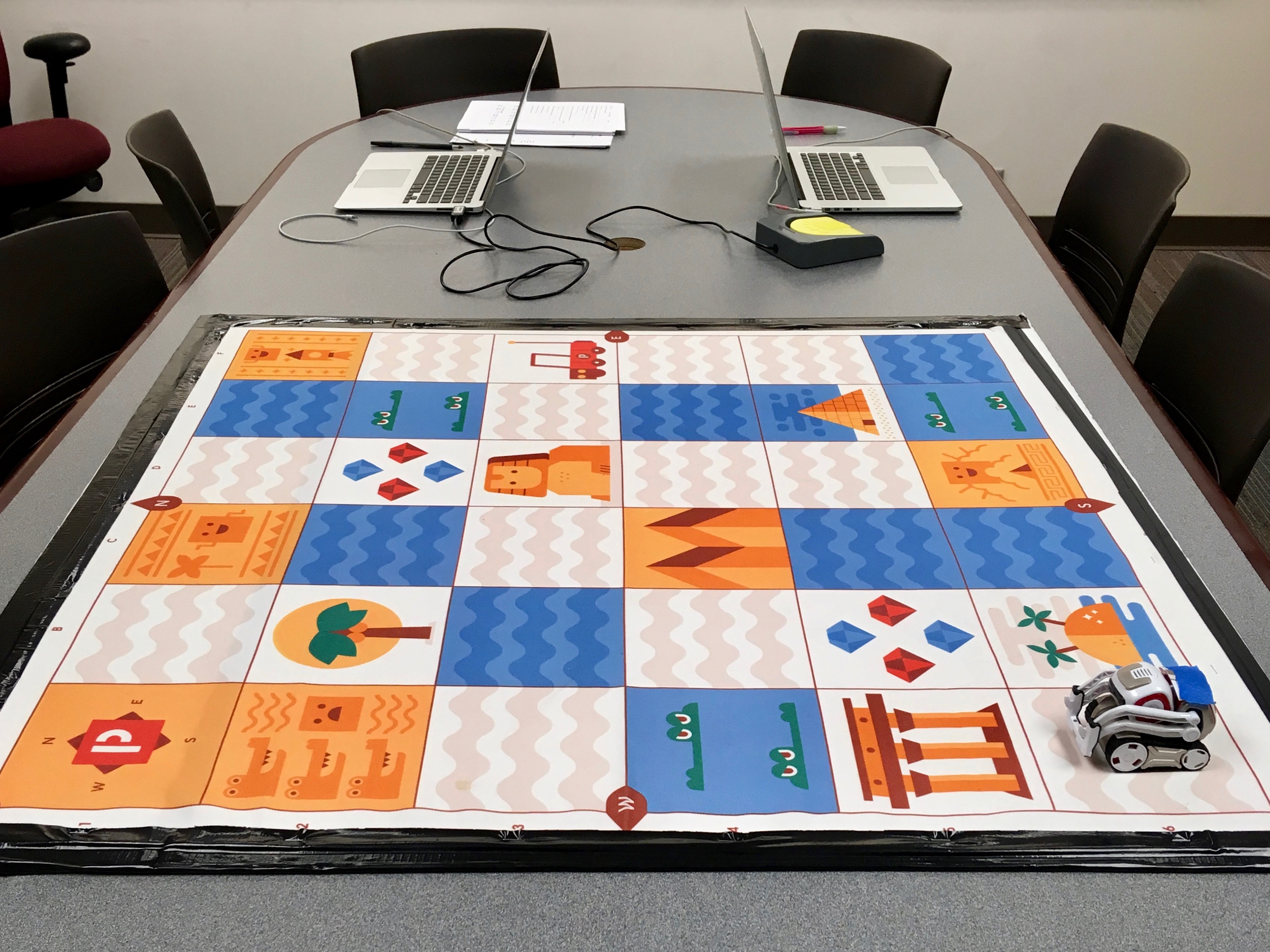}
\label{fig:snake}
}
\subfigure[Snake Game]{
\includegraphics[width=2.5in]{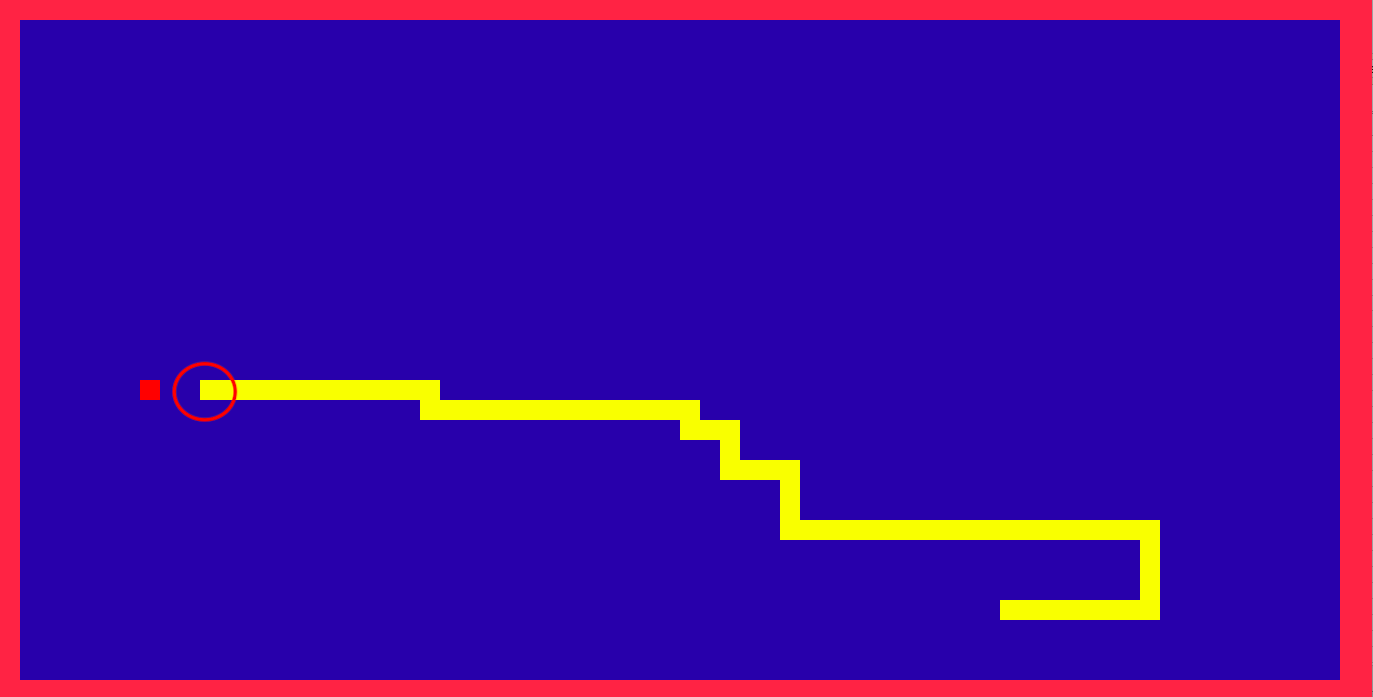}
\label{fig:room}
}
\caption{(a) The experimental setup shows the robot on the Egypt map, the participant computer for the online Snake game, and the experimenter's computer logging the robot's behavior and the button presses from the yellow button. (b) Participants were asked to play the Snake game by pressing the arrow keys to move the snake head (indicated with a red circle) over the red food pieces while avoiding hitting itself and the red walls around the board.}
\end{figure*}

The experiment took place in a small conference room with an oblong table about 1.3 by 3.5 meters in size. On one half of the table were two places for people to sit facing each other, one for the experimenter and the other for the participant. A laptop was positioned at each spot, and a USB-linked button was positioned to the left of the participant laptop and connected to the experimenter laptop. The other half of the table was used for the three maps, each of which had been affixed to a piece of foam core in order to ensure that it would stay flat enough for the robot to traverse. Before each trial, the experimenter placed the appropriate map to the left of the participant and positioned the robot in the correct square. The setup is shown in Figure~\ref{fig:room}.

\subsubsection{Conditions}
All of the participants saw each of the three different path conditions (No Error, Early Error, Late Error), one on each of the three maps for the within-subjects variable Error Finding. They saw the Egypt map first, followed by the Space map and the Street map. Map order was held constant because of technological constraints. The order of the three Error Finding conditions (No, Early, or Late Error) was randomized for each participant (six total combinations). Alternating participants were assigned to one of the two Explanation conditions: Explanation or No Explanation. 

\subsection{Participants}
Participants were recruited using a community research recruitment website run by the university. In order to take part in this research, participants had confirm that they were 18 years of age or older and had normal or corrected-to-normal hearing and vision. Sixty individuals successfully completed the experiment (29/30/1 female/male/nonbinary; age range 19-61 years, \textit{M} age = 28.65, \textit{SD} age = 10.39), including five in each of the twelve combinations of conditions (6 Error Finding x 2 Explanation). They provided informed consent and received compensation for their time. This research was approved by our Institutional Review Board. 

\subsection{Procedure}
Upon arrival at the lab, each participant provided informed consent and was given the opportunity to ask the experimenter questions. They then completed a questionnaire about demographics (including age, gender, languages spoken, country of origin, field of study, and familiarity with robots, computers, and pets) and the Ten-Item Personality Inventory \cite{gosling2003very}. 

The participant was then told that the goal of the experiment was to assess people's ability to simultaneously monitor the robot while completing their own task. The experimenter introduced the Snake game and the participant was given the opportunity to practice playing Snake on the laptop for up to five minutes (as long as it took for them to feel comfortable) in the absence of any other task. Next, the experimenter instructed the participant that there would be three scenarios in which the participant would play Snake as much and as well as possible while also monitoring the robot as it completed its map navigation task. The participant was told to press the yellow button to the left of the laptop when the robot entered the indicated squares and that the button would make the computer beep to record the feedback, but the robot would continue entering the square. The participant was asked to press the button for familiarization and to ensure firm presses. 

The experimenter set up the Cozmo robot and the first map. She told the participant that Cozmo would be navigating the map and to press the button if it ventured into the relevant squares. The participants in the Explanation condition were told specifically that the Cozmo had been programmed to avoid these squares and to press the button if it entered them anyway. Participants in the No Explanation condition were told to hit the button if Cozmo entered specific squares. For each participant, the experimenter selected a random order of Error Finding conditions and the robot was prepared to complete the first condition. The experimenter and participant verbally coordinated so that the Snake game and the robot navigation began simultaneously. After approximately one minute (range 57-67 seconds), the Cozmo completed its journey and the experimenter instructed the participant to end the Snake game (i.e., let the snake crash into the wall). The participant then completed a survey about their trust in the robot and their ability to complete the two simultaneous tasks. The same procedure was then repeated for the second and third maps. For each map, the robot had a piece of colored tape covering its back in order to enable the conceit that three different robots were being used. This tape was switched out of sight of the participants, so it appeared as though the experimenter had brought a different robot to the table. We provided this visual differentiation to attenuate the effects of participants developing mental models of the robot across maps. 

\subsection{Measures}

We used participant performance on the Snake game and their ability to detect robot navigation errors as objective measures. Subjective measures included questionnaire responses from the participant after each trial.

\subsubsection{Snake Game Task Objective Measures}

To analyze performance on the Snake task, we created windows that extended 10 seconds before and after the time at which the robot was programmed to commit an early or late error for each map. We created three variables: \textit{key count}, the number of times a participant pressed a key to control the game during the 20-second window; \textit{key rate}, the average time between each key press, measured in milliseconds; and \textit{death count}, the number of times the participant died in the game during the window. We were thus able to compare behavior across the two 20-second windows for each map and determine the degree to which game performance was affected by the occurrence of an error in one specific window (errors only occurred in one of the two windows per map). We used these data as proxy measures for participant attention to the game at any given time and examined how these numbers corresponded to the status of the robot and the experiment condition.


\subsubsection{Robot Monitoring Task Objective Measures}

Using the Cozmo log files, we calculated the latency between Cozmo entering an error square and participant button press to notify us of the error. These response times were compared across conditions to determine how the timing of an error and the task explanation affected participant performance on the Error Finding task. We also noted if the participant neglected to report any errors that did occur.

\subsubsection{Subjective Measures}
Participants completed a questionnaire after every trial of the study that included 15 rating questions and a question about estimating the number of errors made by the robot. The rating questions were completed on a 7-point scale ranging from Strongly Disagree to Strongly Agree and included questions on wariness, confidence, robot dependability, robot reliability, robot predictability, the extent to which the robot could be counted on to do its job, the degree to which the robot was malfunctioning, participant trust in this robot, participant trust in all robots generally, whether the participant will trust robots as much as before, whether the robot made a lot of errors, whether the participant could focus on Snake or if the robot required too much attention, whether the participant spent more time on Snake or robot monitoring, whether it was hard to complete the Snake task during robot monitoring, and whether the participant would spend more time watching the robot if doing the study again. Many of the questions on the post-experiment questionnaire were adapted from previous research by Jian and colleagues \cite{jian2000foundations} and Muir \cite{Muirthesis}; others were created specifically by us to assess dual-task experiences.

\subsection{Hypotheses}
We hypothesized that the explanation of the robot's behavior allows participants to anticipate the robot's behavior so that they can be more selective in how they focus their attention between the two tasks. In terms of our measures, we expected explanations to result in better game task performance (higher key counts, lower key rate, and fewer snake deaths) compared to no explanations (H1). We thought that the explanations would have a greater effect when the robot is novel and diminish over time (H2), and they would lead to higher subjective measures of trust (H3). Additionally, following prior work, we expected to find that robot errors reduced both objective and subjective trust measures (H4). 


%% file: Results.tex
\section{Results}

We used performance metrics from the two tasks in the experiment and responses to the questionnaires to assess trust in the robot both directly and indirectly. 

\begin{figure*}[t]
    \centering
    \subfigure[]{
    \label{fig:key_error}
    \includegraphics[width=0.3\textwidth]{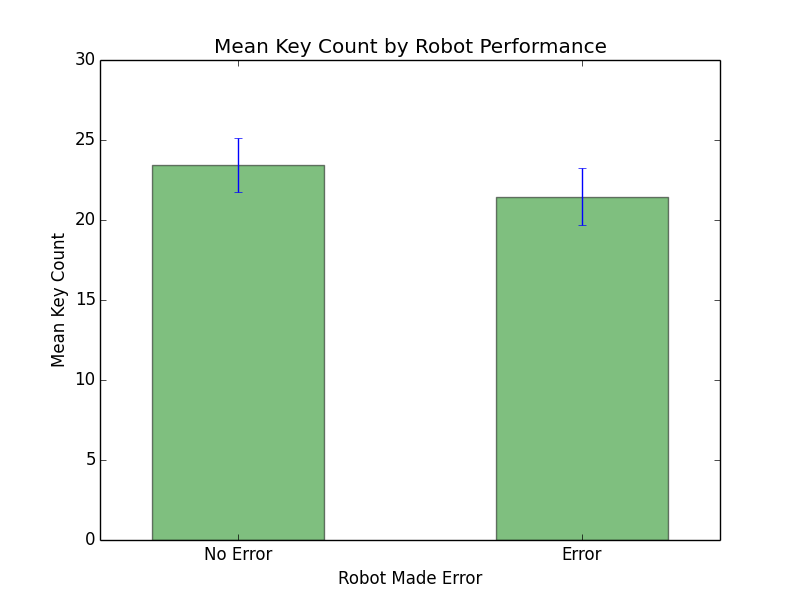}
    }
    \subfigure[]{
        \label{fig:key_exp}
    \includegraphics[width=0.3\textwidth]{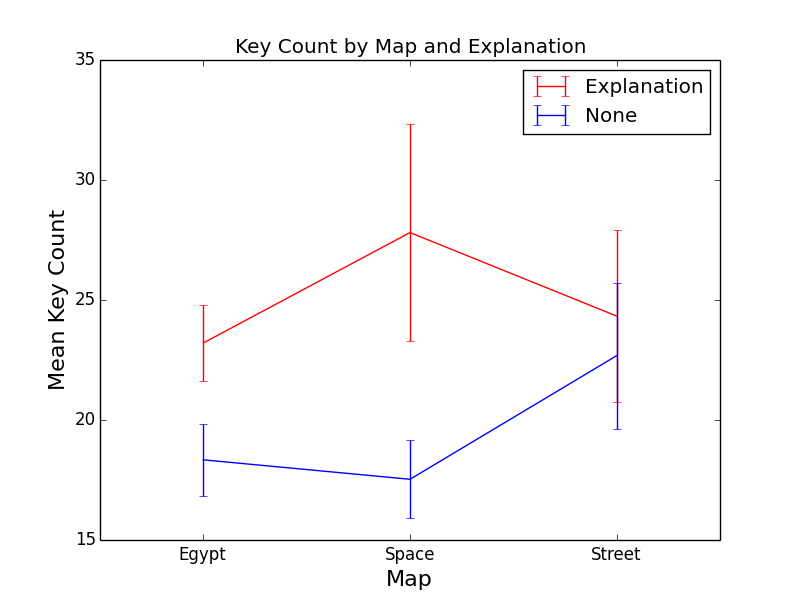}
    }
    \subfigure[]{
        \label{fig:death_exp}
    \includegraphics[width=0.3\textwidth]{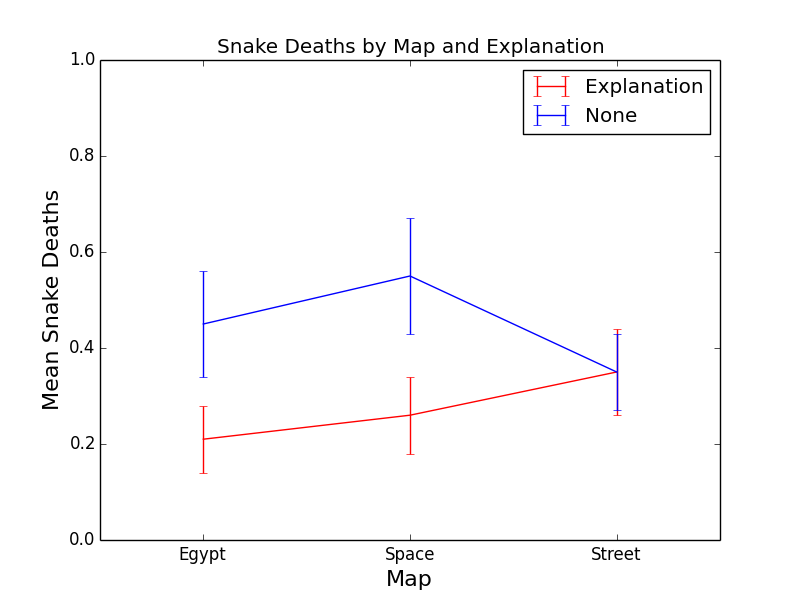}
    }
    \caption{(a) Participants' key counts when the robot was not making an error compared to when it was. (b) Participants who received an explanation made significantly more key presses in the first two maps compared to those who did not. There was no difference between explanation conditions on the last map. (c) Similarly, participants who received an explanation died in the Snake game less frequently in the first two maps, but not the third. }
        \label{fig:results}
\end{figure*}

\subsection{Dual-task performance}

First, we examined \textit{key count} for the Snake game. Participants did not significantly press fewer buttons during time windows in which the robot made an error, \textit{F} = 3.112, \textit{p} = 0.080 (Figure \ref{fig:key_error}). There were no significant main effects of explanation condition, error order, or map. We found a significant interaction between map and explanation condition, \textit{F} = 3.161, \textit{p} = 0.045, such that participants in the explanation condition had higher key counts than those in the no explanation condition for the first two maps, but similar key counts in the last map, although the pairwise comparisons were not quite significant (Figure \ref{fig:key_exp}).

We found a significant main effect on \textit{key rate} for whether there was an error, \textit{F} = 4.868, \textit{p} = 0.029, such that the time between key presses was higher (i.e., a lower key press rate) when the robot made an error than when it did not. No other significant main effects or interactions were found.

For \textit{death count}, there were no significant main effects, but there was a significant interaction between map and explanation condition, \textit{F} = 4.374, \textit{p} = 0.0139 (Figure \ref{fig:death_exp}). Although pairwise comparisons were again not significant, a pattern of effects was found that participants who received no explanation had higher death counts for the first two maps than those who received explanations, but this difference diminished by the third map. There was also a significant interaction between error order and whether there was an error, \textit{F} = 5.536, \textit{p} = 0.0198. An early error with no explanation was most likely to result in death, followed by early error with explanation, late error with explanation, and late error with no explanation. Pairwise comparisons were significant between early error/no explanation and late error/no explanation only. 

We also examined \textit{button press} data to assess whether participants were attending to the robot as it traversed the maps. There were no significant main effects of any of our condition manipulations on how long it took participants to hit the yellow button after the robot entered one of the error squares. In general, participants were very fast and accurate in pressing the button to report robot errors.

Overall, these dual-task performance results suggest that errors made by the robot significantly affected the Snake game task \textit{key rate} and although did not quite affect \textit{key count}, partially in line with our fourth hypothesis (H4) for the objective measures of trust. This result suggests that the supervisory task did require that participants slow down their game performance to report errors for the robot. Participants were able to slow down the key rate without reducing the accuracy of reporting errors and without an increased Snake death count. Additionally, although our findings do not support our hypothesis that explanations would improve gameplay overall (H1), the Explanation condition had notable effects on key count and death count in the beginning of the experiment on the first two maps and decreased for the last map, providing some support for hypothesis H2.

\subsection{Questionnaires}

\begin{table}[]
    \centering
    \begin{tabular}{l|l}
        Questionnaire Item & Significant Effects  \\
        \hline
        Wary & Error**\\ 
        Confident & Error**\\
        Dependable & Error**\\
        Reliable & Error**\\
        Count on this robot & Error**\\
        Trust this robot & Error**\\
        Predictable & Interaction Map x Error*\\
        Malfunctioning & Error*, Interaction Map x Error*\\
        Trust robots in general & ---\\
        Not trust robots as much & Interaction Map x Explanation*\\
        The robot made errors & Error**\\
    \end{tabular}
    \caption{Significant main effects and interactions for trust-related questionnaire items. * = \textit{p} $<$ 0.05; ** = \textit{p} $<$ 0.005.}
    \label{tab:my_label}
\end{table}

Participants answered 16 questions after each trial to examine their feelings about the specific robot they had just seen as well as robots in general. For many of these questions, there was a significant main effect of which error condition they had just seen on the participants' responses. 

Ratings of ``I am wary of the robot'' were significantly affected by error condition, \textit{F} = 6.260, \textit{p} = 0.003, such that ratings for early error and late error were significantly higher (measured by Tukey HSD pairwise comparisons) than ratings when there was no error, \textit{p} $<$ 0.05. There was not a significant main effect of explanation condition, \textit{F} = 3.471, \textit{p} = 0.068. 

Similarly, there was a significant main effect of error condition for ``I am confident in the robot,'' \textit{F} = 10.628, \textit{p} $<$ 0.0001, such that the ratings for no error were significantly higher than early and late error, p $<$ 0.05. The same pattern for error condition, including pairwise comparisons, was observed for ``The robot is dependable,'' ``The robot is reliable,'' ``To what degree can you count on the robot to do its job?'', and ``I trust this robot,'' all \textit{F} $>$ 12.000, all \textit{p} $<$ 0.0001, all pairwise comparisons significant for no error versus early and late errors \textit{p} $<$ 0.05. For these questions, there were no other significant main effects or interactions. 

For the item ``To what extent can the robot's behavior be predicted from moment to moment?'', there were no significant main effects of explanation, map, or error condition. There was a significant interaction between map and error condition, \textit{F} = 3.118, \textit{p} = 0.017, but no significant pairwise comparisons were identified using Tukey HSD. 

Ratings of ``The robot was malfunctioning'' were significantly affected by error condition, \textit{F} = 11.448, \textit{p} $<$ 0.0001. Pairwise comparisons found that these ratings were significantly lower for the no error condition (\textit{M} = 1.517, \textit{SD} = 0.965) than for the early (\textit{M} = 2.183, \textit{SD} = 1.432) or late (\textit{M} = 2.267, \textit{SD} = 1.388) error conditions, \textit{p} $<$ 0.05. There was also a significant interaction between explanation and error condition, \textit{F} = 3.205, \textit{p} = 0.045, such that explanations combined with early and late errors elicited significantly higher ratings than when there were no errors, regardless of explanation condition. Having no explanation combined with either early or late error produced intermediate ratings that were not significantly different from other combinations' ratings. The explanation says that the robot is programmed to avoid those squares, resulting in an assessment of malfunction when it does. 

None of our manipulations affected ratings of ``I trust robots in general.'' There were no significant main effects of our manipulations on ratings for ``I will not trust robots as much as I did before,'' although there was an interaction between map and condition, \textit{F} = 2.550, \textit{p} = 0.0416. No pairwise comparisons were significant, however. These two questions sought to measure whether our study affected trust in robots beyond the experiment itself.

We asked participants two questions specifically about how many errors the robots made. For ``The robot made a lot of errors,'' there was a significant main effect of error condition, \textit{F} = 23.093, \textit{p} $<$ 0.0001, such that early and late errors elicited significantly higher ratings than when there was no error, \textit{p} $<$ 0.05. When prompted to ``Estimate the number of errors made'', participant ratings had a main effect of error condition, \textit{F} = 149.079, \textit{p} $<$ 0.0001. They were generally extremely accurate in identifying the number of errors made. 

We asked participants a few questions about their dual-task experience. There were no significant main or interaction effects for ratings of ``I could not focus on the Snake task because the robot needed my attention.'' For ratings of ``I spent more time watching the robot than on the Snake task,'' there was only a significant interaction of error condition and explanation, \textit{F} = 3.239, \textit{p} = 0.0431, such that an early error with an explanation elicited higher ratings than no error with an explanation (\textit{p} $<$ 0.05), but no other pairwise comparisons were significant. Ratings for ``It was hard to complete the Snake task while watching the robot'' showed only a significant main effect of which map was being used, \textit{F} = 4.161, \textit{p} = 0.0182, with ratings for the first map being higher than the second and third maps (\textit{p} $<$ 0.05), suggesting that perceived difficulty decreased with practice. Finally, ratings for ``If I did this study again, I would spend more time watching the robot'' was not significant for error condition, \textit{F} = 2.821, \textit{p} = 0.0641, with not-quite-significantly higher ratings for early error than for late error or no error. There was a significant interaction of map and error condition, \textit{F} = 2.568, \textit{p} = 0.0407, but no significant pairwise comparisons. 

Overall, the questionnaire responses clearly reflect that participants were monitoring the robot's performance levels, and errors made by the robot were reflected in assessments including trust, dependability, and reliability. These findings provide partial support for our fourth hypothesis (H4) by confirming that errors reduced subjective measures of trust. Having an explanation for the robot's behavior had no major, independent effects on questionnaire responses. This fails to confirm our hypothesis H3 that explanations would improve subjective measures of trust.  


%% file: Discussion.tex
\section{Discussion}

Our results partially supported our hypothesis H2 that explanations of the robot's behavior would significantly affect the participants' gameplay during early trials of the dual-task experiment but not in the last trial, when the robot was more familiar. By the third trial, the participants who received no explanation for the robot's behavior improved their gameplay enough that the explanation did not matter. However, there was no main effect of explanations on objective trust (H1) nor subjective trust (H3) throughout the entire experiment. Additionally, there was some support for our hypothesis H4 that participant trust, measured both by gameplay and by questionnaire, was significantly affected by the robot's errors. 

\textit{Role of Explanations.} Neglect tolerance measures in our dual-task experiment suggest that errors in robot performance deflect effort from the game task to increase monitoring of the robot. While robot errors reduced participant neglect tolerance (supporting H4), providing explanations for the robot's behavior boosted this tolerance during early trials (supporting H2). We provided a relatively simple explanation for the robot's task: it was programmed to avoid certain squares. Alternatively, participants with no explanation were simply told to hit the button when the robot entered those squares. While the explanation was not long nor very specific about the robot's path, it still significantly impacted the participants in the task. It is possible that the explanation led participants to maintain their focus on the game rather than spending more effort tracking the robot's movements because it suggested that the robot ought not enter those squares and would actively avoid them. It is likely that this impact on neglect tolerance was higher when the situation was still novel because the participants had not seen very many errors occur at that point and had not created their own updated mental models for the robot's performance. 

Additionally, providing different types of explanations for robot behavior could also change neglect tolerance. Our explanation suggested that the robot would avoid entering certain areas of the map, which could bias the observer's mental model to assume that the robot would not make errors. Alternative explanations, including which landmarks the robot passes over or what turns it makes through the map, could bias the person further in the same direction by providing more detail about the robot's programming and/or emphasizing that entry to those areas is a mistake, or they could bias the person to think it is not particularly important whether the robot enters those areas. It is possible that the effects of any explanations would be attenuated by a more challenging task competing with supervision of the robot. Future research should examine the effects of multiple levels of explanations and task difficulty on neglect tolerance. 


\textit{Subjective Ratings of Trust.} As predicted in hypothesis H4, the presence versus absence of an error had significant negative effects on many participant ratings of the robot, including measures of trust, reliability, and dependability. However, ratings of robot malfunction were generally low even after an error had occurred. Notably, whether participants had received an explanation of robot behaviors did not significantly affect their ratings of the robot (contradicting H3). 

Overall, the questionnaire results did not reflect the changes in behavior that were observed, indicating that subjective measures of trust are not sensitive enough to catch subtle differences for certain tasks. In order to accurately measure robot autonomy and the ability of a person to do another task while still monitoring the robot, questionnaires do not properly evaluate that level and type of trust (as found in \cite{desai2013impact} and \cite{yang2017evaluating}). Recording and assessing data from the dual task provided a better measure of trust through neglect tolerance.


\textit{Dual Task Experiment Design.} Our task was brief and each trial included no more than one error. To learn more about how people allocate attention and effort, future research should investigate the effects on neglect tolerance of different robot error rates and amounts. Frequent errors or near-misses might close the gap between observers who did and did not receive explanations because it would quickly force reassessment of the observers' mental models. Moreover, an increase in these factors would likely result in worse performance on the other task. Additionally, attention and effort allocation could be biased towards the alternative task by increasing the difficulty of that task. For our game, it was possible to slow down the button presses and avoid hitting obstacles in order to avoid losing the game; however, a game with more obstacles or opportunities to win points in shorter time spans might elicit more effort from the player and divert attention away from the robot. For real-world robot supervision, it is important to know what task is appropriate for people to do in addition to noticing robot behaviors. 

\textit{Novelty Effect.} Finally, our examination of novelty was relatively limited. An increase in the number, variety, and length of trials would allow further assessment of the degree to which explanations matter as someone gains more experience with the robot. Moreover, it is possible that map order impacted our results. There also are likely long-term effects of practice on both tasks. Novelty effects might also relate to task difficulty such that explanations impact user's mental models about the robot for a longer period of time if they are expending their effort on the other task because they do not have the cognitive effort available to update these models. 





%% file: Conclusion.tex
\section{Conclusion}
We conducted a dual-task experiment to study the effect of explanations on robot trust. We measured participants' neglect tolerance---the time that participants spent watching our robot versus performing their own task---as well as subjective trust through surveys. While explanations did not have a main effect on objective or subjective trust measures, they did have an effect that counteracts the novelty of seeing a new robot for the first time. Additionally, we found that our neglect tolerance measure was able to identify subtle changes in trust compared to survey measures that did not find significant differences across conditions in the study.
